\documentclass[11pt]{article}

\usepackage[preprint]{acl}

\usepackage{times}
\usepackage{latexsym}

\usepackage[T1]{fontenc}

\usepackage{microtype}

\usepackage{inconsolata}

\usepackage{graphicx}
\usepackage{float}
\usepackage{svg}

\usepackage{soul}

\usepackage{listings}

\title{LLM-based Visual Code Completion for Aerospace Geometric Design}

\author{
    \textbf{Hau Kit Yong\textsuperscript{1}},
    \textbf{Robert Marsh\textsuperscript{1}},
    \textbf{Edmar A. Silva\textsuperscript{1}},\\
    \textbf{András Sóbester\textsuperscript{1}},
    \textbf{Stuart E. Middleton\textsuperscript{2}} \\
    \textsuperscript{1}Faculty of Engineering and Physical Sciences, University of Southampton, UK \\
    \textsuperscript{2}School of Electronics and Computer Science, University of Southampton, UK \\
}

\begin{document}
\maketitle
\begin{abstract}
Recent advances in both Large Language Models (LLMs) and Vision Language Models (VLMs) have seen a step change in their ability to perform visual code completion, but the aerospace industry, which prioritizes safety and explainabilty over rapid LLM adoption, currently has no publicly announced LLM-based geometric design copilot systems in commercial use by aerospace Original Equipment Manufacturers (OEMs). This paper presents a LLM-based visual programming copilot application for aerospace engineering design tasks, using a visual programming variant of the ReAct methodology and GPT 5.4. In addition to the copilot, we describe Wingbuilder, a new Grasshopper plugin library with custom components for aerospace-specific geometry abstraction, and an associated Aerospace Visual Programming Dataset (AVPD) with 18 aerospace expert designed tasks at different levels of difficulty alongside ground truth solutions. We evaluate our copilot application with a user trial involving two experienced aerospace engineers from a large aircraft manufacturing company. We find our copilot visual programming ReAct methodology was successful in generating suggestions that participants found helpful, but slow ReAct inference times limit its usefulness to more complex time-consuming tasks where waiting for good copilot solution suggestion was worthwhile. Participants reported they liked the tool and would be willing to use it in the future.
\end{abstract}

\section{Introduction}
Visual programming environments support many application domains, and are typified by systems such as LabView \citep{Kodosky2020LabView} and Visual Blocks \citep{Du2023Experiencing}. In aerospace engineering parametric Computer Aided Design (CAD) tools \cite{SobesterForrester2014} \cite{Silva24} use visual programming to create design components that are executed to render 3D models, such as an aircraft's fuselage or wing. Recent advances in both Large Language Models (LLMs) and Vision Language Models (VLMs) have seen a step change in their ability to perform visual code completion, but the aerospace industry, which prioritizes safety and explainabilty over rapid LLM adoption, currently has no publicly announced LLM-based geometric design copilot systems in commercial use by aerospace Original Equipment Manufacturers (OEMs). More scientific evidence is needed of LLM effectiveness and safety before mainstream adoption will happen in the aerospace industry.

In this paper we create and evaluate a LLM-based copilot application for helping aerospace engineers with the visual programming tasks associated with creating new aerospace wing geometry designs. Rhinoceros 3D\footnote{https://www.rhino3d.com/} is a Non-Uniform Rational Basis Splines (NURBS) based 3D modeller widely used in the engineering domain for architectural and industrial design. Grasshopper\footnote{https://www.grasshopper3d.com/} is the built-in visual programming environment for Rhinoceros 3D. We first describe WingBuilder, our new Grasshopper plugin that provides a new palette of custom components for aerospace-specific geometry abstractions, such as airfoils, wing segments and cross-sections. We then describe our LLM-based visual programming copilot application, which recommends to a user Grasshopper canvas edit actions using GPT 5.4 prompted with a screenshot of the current canvas, system prompt description of available Wingbuilder components and a description of the task the engineer wants to achieve. Our LLM-based recommendations are generated using a ReAct-style \citep{yao2023react} methodology, adapted to not include a visual analysis step so that we can exploit better the pre-trained knowledge GPT 5.4 has about Grasshopper visual component layout. We release both our Wingbuilder library\footnote{{REDACTED github URI pending acceptance decision}} and LLM-based copilot application\footnote{{REDACTED github URI pending acceptance decision}} as open source for the aerospace and visual programming community to use in the future.

To evaluate our LLM-based copilot we conduct a task-based user trial with two aerospace engineers from Airbus, where engineers are asked to perform aerospace-specific visual programming tasks of varying difficulty with and without our LLM-based copilot. We evaluate both in-task using quantitative metrics and post-task using qualitative metrics, reporting our experience in the form of both best practice and lessons learnt to guide future researchers in developing aerospace-domain LLM-based applications in the future.

This work makes the following contributions: 
\begin{itemize}
\item We describe a new geometry abstraction library, called Wingbuilder, designed explicitly to support the task of visual programming for aerospace wing design and supporting LLM-based automation.
\item We describe a new visual programming task focussing on authentic  aerospace wing design and associated Aerospace Visual Programming Dataset (AVPD) with 18 aerospace expert designed tasks tailored for evaluating LLM-based geometric visual code completion. 
\item We report on insights gained developing and evaluating a LLM-based copilot application for aerospace-based visual visual programming. We provide quantitative and qualitative evaluation results from a user trial involving two aerospace engineers from a major aerospace company, and discuss best practice to guide future researchers in this area.
\end{itemize}

\section{Related Work}
\subsection{LLMs for visual programming}

LLMs have been widely used to support visual programming environments, which typically feature workflows or node/block graphs with examples including LabView \citep{Kodosky2020LabView}, Visual Blocks \citep{Du2023Experiencing} and ComfyUI \citep{huang2025comfyuicopilot}. Recent work on LLM-assisted visual programming has focussed on agentic LLM frameworks, LLM distillation, LLM preference optimisation and the use of various visual cues as prompts to support visual code generation tasks.

An example of a LLM-based agentic framework for visual programming is \citep{xue2025comfyuir1}, where Chain-of-Thought (CoT) and reinforcement learning is used to automate visual programming workflow generation for image creation tasks. Agentic LLM planning has also been used to format Simulink visual programmes \citep{simuagent2026}, using a reflective Group Relative Policy Optimization (GRPO) stratgey and generate pseudocode for node-graph workflows for machine learning pipeline tasks \citep{zhou2025instructpipe}. Non-agentic approaches include Vison Language Model (VLM) distillation \citep{shlapentokh-rothman-etal-2025-visual} and VLM preference optimization \citep{kang-etal-2025-retrieval} for visual programming tasks such as working with logic-based ladder diagrams. Visual cues \citep{zhao2024chatscratch} have been used with VLMs as prompts for block programming code completion environments such as Scratch\footnote{https://scratch.mit.edu/} alongside standard evaluation frameworks such as ScratchEval \citep{fu-etal-2025-scratcheval}.

\subsection{LLMs for parametric Computer Aided Design (CAD)}

In parametric CAD, the artefacts of interest are the CAD operators that create things like extrudes and fillets in a model. Analagous to software code being executed to achieve a goal, parametic CAD can be rendered to generate a 3D geometric design. Work on LLM-assisted CAD can be broken down into two main approaches, one-shot CAD generation and iterative CAD agents.

One-shot generation creates a set of CAD instructions when prompted with a design task. Recent work includes Transformer-based autoencoders \citep{wu2021deepcad}, LLMs augmented with VLMs such as LLaVA-NeXT \citep{liu2024llavanext} to create shape descriptions \citep{khan2024text2cad}, data synthesis to convert CAD sequences to code more suitable for LLMs \citep{li2025cadllama} and use of LLMs to generate CAD instructions from image-based prompts such as sketches and images \citep{wu2024CadVLM}.

CAD agents methods introduce agentic planning and reasoning steps, iteratively generating CAD instructions. Recent work include use of VLMs to plan/reason and then generate python code to execute CAD operations \citep{mallis2025cadassistant}, agentic LLM workflows for planning CAD operations and visual inspection of results for plan improvement \citep{cambridge2024fromtexttodesign} and ReAct-style LLM reasoning to plan CAD operations \citep{caddesigner2025} \citep{yao2023react}. The CAD community is starting to adopt some of these LLM-based approaches in AI-plugin assembly projects such as Raven\footnote{https://www.food4rhino.com/en/app/raven}, which provides a user interface on top of LLM-based support tools for engineering and architectural design tasks much simpler than the complex aerospace wing design we address in this paper.

Our approach is motivated by the ReAct-style LLM-based reasoning approach of \citep{caddesigner2025}, but unlike this work we do not render intermediate models for LLM visual analysis, allowing us to work directly with the visual programming canvas and therefore exploit pre-trained LLM coding knowledge much better. We also test our method on aerospace wing design tasks, much more complex in nature than an individual engineering component design tasks like making a flange and or screw, which are used in current LLM-based CAD papers.

\section{Aerospace Visual Programming Dataset and Wingbuilder Library}
\label{sec:dataset}

To support aerospace-domain geometric visual programming in the Grasshopper parametric CAD environment we created Wingbuilder\footnote{REDACTED github URI pending acceptance decision}, the first geometry abstraction library focussing specifically on the challenging task of aircraft wing design.

Wingbuilder has 54 custom Grasshopper components grouped into seven domain categories: 10 for airfoil construction and transformation; 8 for wing assembly (planform construction, segment lofting, wing building); 14 part factories, covering common structural elements and control surfaces; 8 cross-section primitives for assembling custom parts; 8 analysis modules; 5 geometry evaluation utilities; and one custom dimensional parameter for units-aware inputs. By exposing the wing-design workflow as a small set of higher-level domain objects (airfoil, wing segment, spar, rib, and so on) rather than the underlying Grasshopper geometric primitives, Wingbuilder lets engineers think and build in their native vocabulary, with each component's inputs labelled in aerospace domain terms (chord, sweep, dihedral, span position) rather than abstract geometric coordinates. Using Wingbuilder, tasks such as adding a structural spar pattern, applying a linear washout schedule across both segments of a multi-segment wing, or running a mass analysis on a parametric design are reduced from creating dozens of low-level Grasshopper primitives to a handful of Wingbuilder components.

Alongside our Wingbuilder library we have created an Aerospace Visual Programming Dataset (AVPD) containing 18 aerospace Grasshopper canvases using custom Wingbuilder components, designed by a postdoc-level aerospace expert with an average of 37 components per canvas. Each canvas has an associated aerospace visual code completion design task with a difficulty level of either easy (2 minute task), medium (4 minute task) or hard (8 minute task). A free text task description is included with each canvas, alongside a fully completed canvas, the gold answer, and a version of the canvas with some components hidden, which is the starting point of the task. The dataset also includes textual descriptions of the wingbuilder custom Grasshopper components designed to be included in code completion task prompts for GPT-5.4. Examples of the task canvases and the rendered 3D wing designs they produce can be seen in appendix~\ref{sec:appendix_designs}.

\section{Visual Programming Copilot}

Our approach to making a LLM copilot for aerospace geometry visual programming was motivated by the ReAct \citep{yao2023react} methodology which uses LLMs for \textbf{Re}asoning and \textbf{Act}ions. Extending the classic Chain of Thought methodology, the original ReAct methodology uses a LLM to iteratively generate reasoning thoughts, track and update action plans, and suggest actions to gather additional information from external sources such as knowledge bases. Our visual programming copilot tailors the text-based ReAct method to visual programming, reasoning about components in a visual programming canvas. Differentiating our approach from approaches such as \citep{caddesigner2025}, we formulate our visual programming ReAct methodology without any visual analysis steps, allowing us to fully exploit the pre-trained knowledge GPT 5.4 already has about Grasshopper visual component layouts.

\begin{figure*}[t]
    \centering    \includegraphics[width=\textwidth]{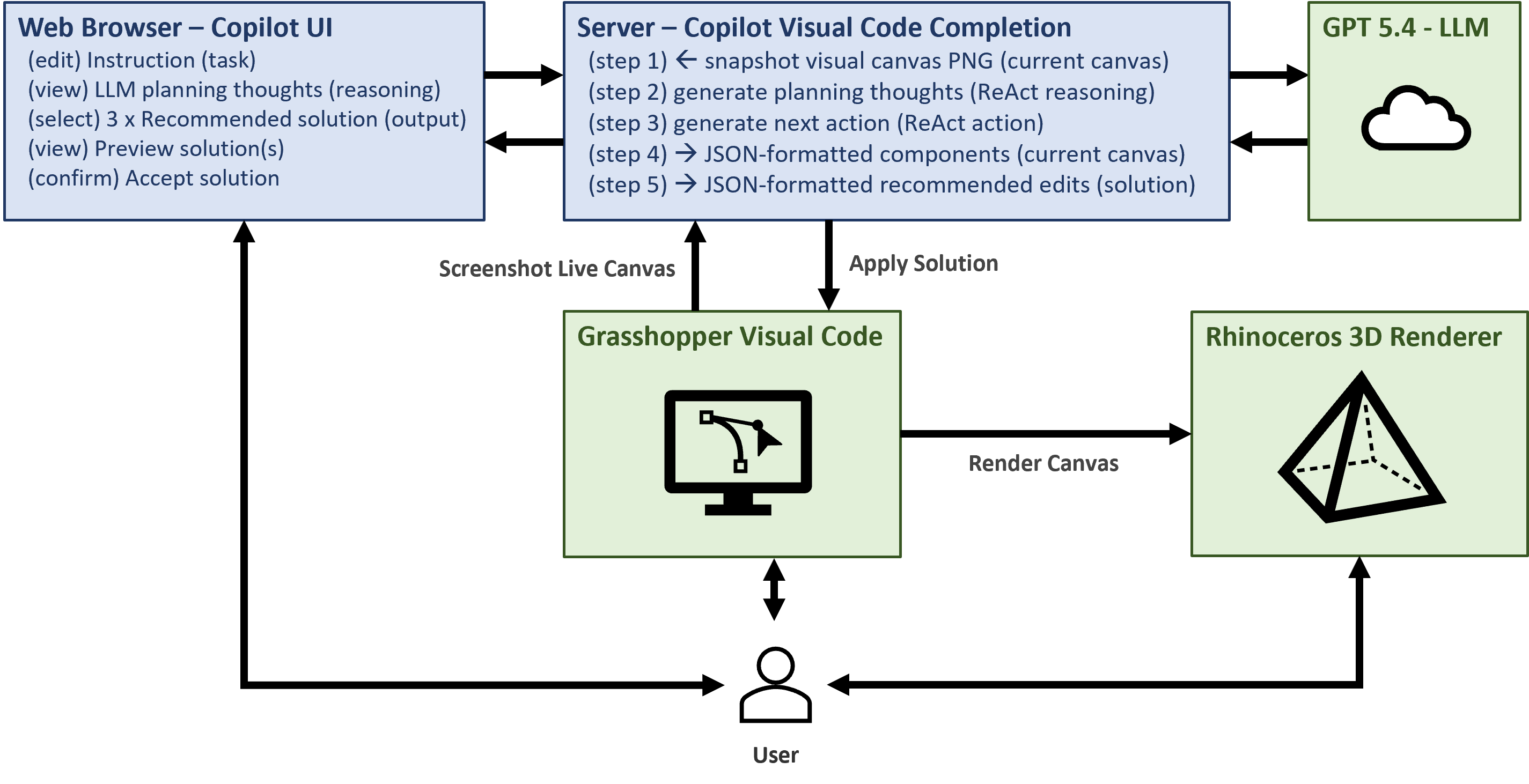}
    \caption{LLM-based Visual Copilot Architecture. Users ask the copilot to suggest component edit solutions to wing design tasks via a ChatGPT style web-interface. Recommended solutions can be previewed in Grasshopper, accepted and added to live canvas, or rejected. Manual edits can be made at any time and the copilot will adapt. A wing design session continues until the aerospace engineer is satisfied with the results.}
    \label{fig:architecture}
\end{figure*}

\begin{figure*}[t]
    \centering    \includegraphics[width=\textwidth]{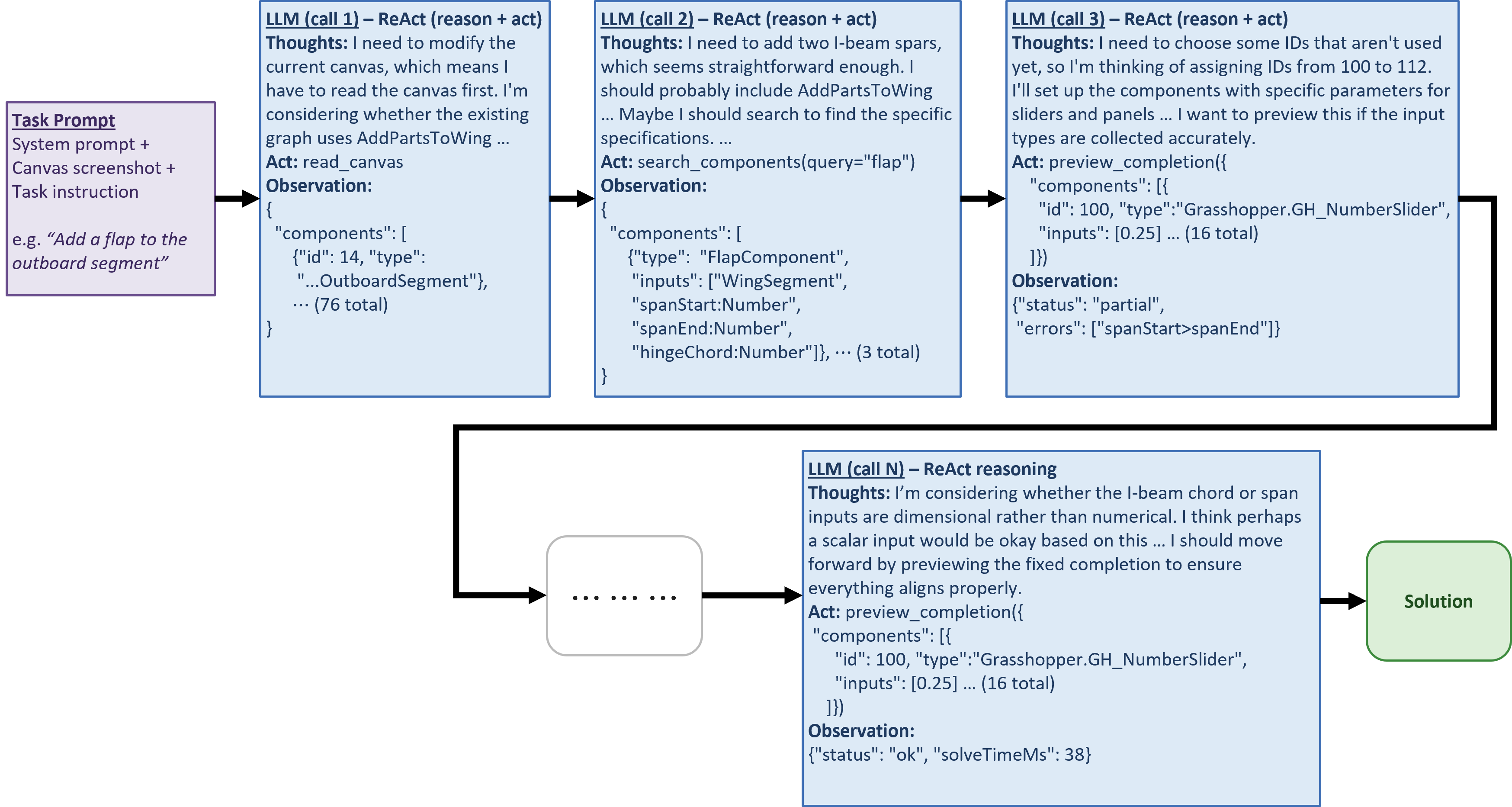}
    \caption{ReAct prompt workflow showing the GPT 5.4 reasoning steps when calculating a visual programming component edit solution to a task.}
    \label{fig:react_workflow}
\end{figure*}

Figure~\ref{fig:architecture} shows our copilot architecture and figure~\ref{fig:react_workflow} the ReAct workflow we use for visual programming. With each ReAct workflow call, the LLM is prompted to first generate free-text thoughts regarding what actions might be needed to achieve the task, and second identify a specific action to perform next. This action is then executed using a set of Grasshopper scripts provided by our copilot server, which results in a set of observations, either suggested new component options or a report on the status of the current canvas. The LLM then uses these observations to generate thoughts as part of the next ReAct workflow call, repeating until the LLM considers the task complete and ready for the human review step.

As can be seen in figure~\ref{fig:react_workflow}, the LLM thoughts are provided to the engineers via the copilot UI to help explain why components are being recommended. At anytime the aerospace engineers can edit the visual canvas, and the copilot will adapt to the new component edits making it an optional tool that recommends new component options to complete the task that the engineers can choose to preview, use or discard. Examples of the LLM prompts we use and the reason/act output traces they generate are provided in appendix~\ref{sec:appendix_prompts}. 

\section{User Evaluation}

\begin{table}[H]
\centering
\small
\begin{tabular}{|l|l|r|r|}
\hline
\textbf{Task} & \textbf{Avg Time (\# completed)}\\

\hline
Easy difficulty (no copilot)&34s (6 of 6)\\
Easy difficulty (copilot)&78s (6 of 6)\\
Medium difficulty (no copilot)&62s (4 of 6)\\
Medium difficulty (copilot)&172s (4 of 6)\\
Hard difficulty (no copilot)&259s (2 of 6)\\
Hard difficulty (copilot)&173s (5 of 6)\\

\hline
\textbf{LLM ReAct} & \textbf{Average Time/Steps}\\

\hline
LLM ReAct step time& 9s\\
LLM ReAct solution time& 113s\\
LLM ReAct steps per solution&12\\

\hline
\end{tabular}
\caption{Quantitative metrics for performance of the copilot LLM following the ReAct methodology, and performance of participants broken down by task difficulty. For easy tasks the LLM ReAct compute time slowed down participants considerably compared to not using the copilot. For harder tasks, participants were both quicker and has much higher task completion rates as they often gave up unable to complete the task at all.}
\label{tab:quant-results}
\end{table}

\begin{figure*}[t]
    \centering    \includegraphics[trim=6cm 6cm 6cm 6cm, clip, width=\textwidth]{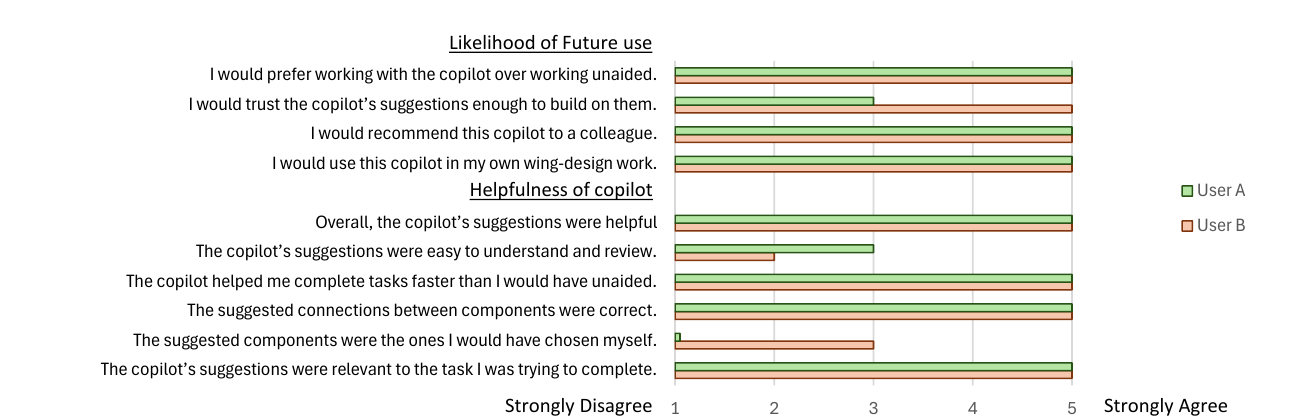} 
    \caption{User trial questionnaire results.}
    \label{fig:user-trial}
\end{figure*}

We evaluated our copilot using a user trial involving two participants, each experienced aerospace engineers at a large aircraft manufacturing company, engaging in a series of aerospace engineering tasks both with and without our copilot tool providing assistance. Based on the Aerospace Visual Programming Dataset described in section~\ref{sec:dataset}, we designed 18 visual programming tasks, each task designed to be either (a) easy difficulty, a two minute task such as adding a single component, (b) medium difficulty, a four minute task requiring more complex component connections, or (c) hard difficulty, an eight minute task involving several components and complex connections. For each task difficulty group we randomly allocated 50\% of tasks to be completed without our copilot tool, the control group, and 50\% of tasks to be completed with our copilot tool available. Tasks were ordered from easy to hard, and always starting with the control group first. Both participants completed the same set of tasks in the same order, which took 120 minutes, and to reduce training bias participants completed a 60 minute training presentation and practice task. If tasks took longer than the two to eight minute task budget they were abandoned and the participant moved on to the next task.

We used a combination of quantitative and qualitative metrics. The quantitative metrics focussed on measuring the copilot LLM ReAct computation time, as this is the most time consuming part of the copilot service provision, the participants speed in completing the tasks and how many tasks were successfully completed. The results can be seen in Table~\ref{tab:quant-results}. Overall the tool showed most benefits for harder tasks, where participants were both faster and more likely to complete the tasks. The LLM ReAct calculations took on average nearly two minutes, which was only really viable for longer tasks. These slow solution calculation times are a reflection on the current state of the art in terms of commercial LLM inference speed, and we would expect this to become faster over time as LLMs become more efficient and the GPU hardware they run on more powerful.

The qualitative metrics were delivered via a post-task questionnaire, completed immediately after the last task was completed. Our questionnaire had six questions about the copilot's helpfulness, four questions about the likelihood of using the copilot in the future and some free text fields for general comments. Responses to questions used a Likert scale (1 = Strongly disagree, 2 = Disagree, 3 = Neither agree nor disagree, 4 = Agree 5 = Strongly agree) and results can be seen in Figure~\ref{fig:user-trial}. Overall participants found the copilot helpful, especially for harder tasks where they struggled to get started with the right solution, but found it did not respond quickly enough to help with easy tasks. The copilot suggestions, whilst useful, were hard to understand and the ReAct reasoning was a little confusing. Participants did express a desire to use such tools in the future, which could suggest a potential paradigm shift is possible for the aeromotive industry with regards LLM digital assistants in the workplace.

\section{Conclusion}

We present a LLM-based visual programming copilot application for aerospace engineering design tasks. We release both Wingbuilder, a new Grasshopper plugin library with custom components for aerospace-specific geometry abstraction, and an associated Aerospace Visual Programming Dataset with 18 aerospace expert designed tasks at different levels of difficulty alongside ground truth solutions. To evaluate our copilot we run a user trial with two experienced aerospace engineers from a large aircraft manufacturing company, measuring quantitative task metrics such as time to complete solution and qualitative metrics via a post-trial participant questionnaire.

We find our copilot visual programming ReAct methodology was successful in generating suggestions that users found helpful when completing the geometric task solutions. Because ReAct involves multiple steps to reach a suggested solution, the LLM was preferred only for medium and hard difficulty tasks that took more than a couple of minutes to complete. The tool was particularly helpful for hard tasks when engineers were stuck and the LLM was able to make helpful suggestions leading to a significantly better task completion rate. Participants reported they liked the tool and would be willing to use it in the future.

For future work we would like to explore how to overcome the existing low performance of open source LLMs for visual programming tasks compared to the powerful commercial LLMs used in this study. This might include exploring sets of heterogeneous base LLMs, fine-tuned for visual programming tasks and grouped into an ensemble. Our hypothesis is that different base LLMs will exhibit different visual programming failure mode patterns, making an ensemble a potentially strong solution. We will also like to expand on our Aerospace Visual Programming Dataset to support more examples for both evaluation and fine-tuning purposes.

\section*{Limitations}

The work in this paper uses a commercial LLM (GPT 5.4) and we observe from our own informal testing that open source LLMs are not currently able to execute the ReAct methodology with sufficient accuracy to make a viable copilot. However, we do expect over time that open source LLMs will improve and become viable, but it is left as future work to verify this.

\section{Acknowledgements}
The DAWS2 (Development of Advanced Wing Solutions 2) project is supported by the ATI Programme, a  joint Government and industry investment to maintain and grow the UK’s competitive position in civil aerospace design and manufacture. The programme, delivered through a partnership between the Aerospace Technology Institute (ATI), Department for Business, Energy \& Industrial Strategy (BEIS) and Innovate UK, addresses technology, capability and supply chain challenges. The authors acknowledge the IRIDIS High-Performance Computing Facility at the University of Southampton.

\section{Ethical Considerations}
This paper's user trial was approved by the University of Southampton's ethical approval panel, ref number ERGO/FEPS/113669. All information gathered in the trial was done so anonymously in accordance with ethical data minimization principles.

\clearpage 

\bibliography{related-work,custom}

@misc{caddesigner2025,
  author = {Fan, Fengxiao and Ni, Jingzhe and Yin, Xiaolong and Wang, Sirui and Lu, Xingyu and Zou, Qiang and Tong, Ruofeng and Tang, Min and Du, Peng},
  title = {{CADDesigner}: Conceptual Design of {CAD} Models Based on General-Purpose Agent},
  year = {2025},
  eprint = {2508.01031},
  archivePrefix = {arXiv},
  primaryClass = {cs.AI}
}

@inproceedings{cambridge2024fromtexttodesign,
  author = {Sch{\"u}pbach, Adrian and San Miguel, Rafael and Ferchow, Julian and Meboldt, Mirko},
  title = {From Text to Design: A Framework to Leverage {LLM} Agents for Automated {CAD} Generation},
  booktitle = {Proceedings of the Design Society 5: ICED25},
  year = {2025},
  publisher = {Cambridge University Press},
  doi = {10.1017/pds.2025.10203}
}

@inproceedings{huang2025comfyuicopilot,
  author = {Huang, Zheng and Hou, Zihui and others},
  title = {{ComfyUI-Copilot}: An Intelligent Assistant for Automated Workflow Development},
  booktitle = {Proceedings of the 63rd Annual Meeting of the Association for Computational Linguistics: System Demonstrations},
  year = {2025},
  eprint = {2506.05010},
  archivePrefix = {arXiv},
  url = {https://arxiv.org/abs/2506.05010},
  note = {Full author list not available in source notes; verify before submission.}
}

@inproceedings{khan2024text2cad,
  author = {Khan, Mohammad Sadil and Sinha, Sankalp and Sheikh, Talha Uddin and Stricker, Didier and Ali, Sk Aziz and Afzal, Muhammad Zeshan},
  title = {{Text2CAD}: Generating Sequential {CAD} Models from Beginner-to-Expert Level Text Prompts},
  booktitle = {Advances in Neural Information Processing Systems (NeurIPS)},
  year = {2024},
  note = {Spotlight presentation},
  eprint = {2409.17106},
  archivePrefix = {arXiv},
  url = {https://arxiv.org/abs/2409.17106}
}

@inproceedings{li2025cadllama,
  author = {Li, Jiahao and Bai, Xueyang and Chen, Yiming and Zhang, Lingxiao},
  title = {{CAD-Llama}: Leveraging Large Language Models for Computer-Aided Design Parametric 3{D} Model Generation},
  booktitle = {Proceedings of the IEEE/CVF Conference on Computer Vision and Pattern Recognition (CVPR)},
  year = {2025},
  eprint = {2505.04481},
  archivePrefix = {arXiv},
  url = {https://openaccess.thecvf.com/content/CVPR2025/papers/Li_CAD-Llama_Leveraging_Large_Language_Models_for_Computer-Aided_Design_Parametric_3D_CVPR_2025_paper.pdf}
}

@inproceedings{mallis2025cadassistant,
  author = {Mallis, Dimitrios and Karadeniz, Ahmet Serdar and Cherenkova, Kseniya and Dupont, Elona and Kacem, Anis and Oyedotun, Oyebade K. and Aouada, Djamila},
  title = {{CAD-Assistant}: Tool-Augmented {VLLMs} as Generic {CAD} Task Solvers?},
  booktitle = {Proceedings of the IEEE/CVF International Conference on Computer Vision (ICCV)},
  year = {2025},
  eprint = {2412.13810},
  archivePrefix = {arXiv},
  url = {https://arxiv.org/abs/2412.13810}
}

@misc{simuagent2026,
  author = {Liang, Yanchang and Zhao, Xiaowei},
  title = {{SimuAgent}: An {LLM}-Based {Simulink} Modeling Assistant Enhanced with Reinforcement Learning},
  year = {2026},
  eprint = {2601.05187},
  archivePrefix = {arXiv},
  primaryClass = {cs.AI}
}

@inproceedings{wu2021deepcad,
  author = {Wu, Rundi and Xiao, Chang and Zheng, Changxi},
  title = {{DeepCAD}: A Deep Generative Network for Computer-Aided Design Models},
  booktitle = {Proceedings of the IEEE/CVF International Conference on Computer Vision (ICCV)},
  year = {2021},
  eprint = {2105.09492},
  archivePrefix = {arXiv},
  url = {https://arxiv.org/abs/2105.09492}
}

@misc{xue2025comfyuir1,
  author = {Xue, Zhenran and others},
  title = {{ComfyUI-R1}: Exploring Reasoning Models for Workflow Generation},
  year = {2025},
  eprint = {2506.09790},
  archivePrefix = {arXiv},
  primaryClass = {cs.AI},
  url = {https://arxiv.org/abs/2506.09790},
  note = {Full author list not available in source notes; verify before submission.}
}

@inproceedings{yao2023react,
  author = {Yao, Shunyu and Zhao, Jeffrey and Yu, Dian and Du, Nan and Shafran, Izhak and Narasimhan, Karthik and Cao, Yuan},
  title = {{ReAct}: Synergizing Reasoning and Acting in Language Models},
  booktitle = {International Conference on Learning Representations (ICLR)},
  year = {2023},
  eprint = {2210.03629},
  archivePrefix = {arXiv},
  primaryClass = {cs.CL},
  url = {https://arxiv.org/abs/2210.03629}
}

@inproceedings{zhao2024chatscratch,
  author = {Zhao, Jiawei and others},
  title = {{ChatScratch}: An {AI}-Augmented System Toward Autonomous Visual Programming Learning for Children Aged 6-12},
  booktitle = {Proceedings of the 2024 CHI Conference on Human Factors in Computing Systems},
  year = {2024},
  eprint = {2402.04975},
  archivePrefix = {arXiv},
  url = {https://arxiv.org/abs/2402.04975},
  note = {Full author list not available in source notes; verify before submission.}
}

@inproceedings{zhou2025instructpipe,
  author = {Zhou, Zhongyi and others},
  title = {{InstructPipe}: Generating Visual Blocks Pipelines with Human Instructions Using {LLMs}},
  booktitle = {Proceedings of the 2025 CHI Conference on Human Factors in Computing Systems},
  year = {2025},
  doi = {10.1145/3706598.3713905},
  eprint = {2312.09672},
  archivePrefix = {arXiv},
  url = {https://arxiv.org/abs/2312.09672},
  note = {Full author list not available in source notes; verify before submission.}
}

@inproceedings{wu2024CadVLM,
author = {Wu, Sifan and Khasahmadi, Amir Hosein and Katz, Mor and Jayaraman, Pradeep Kumar and Pu, Yewen and Willis, Karl and Liu, Bang},
title = {CadVLM: Bridging Language and Vision in the Generation of Parametric CAD Sketches},
year = {2024},
isbn = {978-3-031-72896-9},
publisher = {Springer-Verlag},
address = {Berlin, Heidelberg},
url = {https://doi.org/10.1007/978-3-031-72897-6_21},
doi = {10.1007/978-3-031-72897-6_21},
booktitle = {Computer Vision – ECCV 2024: 18th European Conference, Milan, Italy, September 29–October 4, 2024, Proceedings, Part LXX},
pages = {368–384},
numpages = {17},
location = {Milan, Italy}
}

@inproceedings{Du2023Experiencing,
  title = {{Experiencing Visual Blocks for ML: Visual Prototyping of AI Pipelines}},
  author = {Du, Ruofei and Li, Na and Jin, Jing and Carney, Michelle and Yuan, Xiuxiu and Wright, Kristen and Sherwood, Mark and Mayes, Jason and Chen, Lin and Jiang, Jun and Zhou, Jingtao and Zhou, Zhongyi and Yu, Ping and Kowdle, Adarsh and Iyengar, Ram and Olwal, Alex},
  booktitle = {Adjunct Proceedings of the 33rd Annual ACM Symposium on User Interface Software and Technology},
  year = {2023},
  publisher = {ACM},
  series = {UIST},
}

@article{Kodosky2020LabView,
author = {Kodosky, Jeffrey},
title = {LabVIEW},
year = {2020},
issue_date = {June 2020},
publisher = {Association for Computing Machinery},
address = {New York, NY, USA},
volume = {4},
number = {HOPL},
url = {https://doi.org/10.1145/3386328},
doi = {10.1145/3386328},
journal = {Proc. ACM Program. Lang.},
month = jun,
articleno = {78},
numpages = {54}
}

@inproceedings{shlapentokh-rothman-etal-2025-visual,
    title = "Visual Program Distillation with Template-Based Augmentation",
    author = "Shlapentokh-Rothman, Michal  and
      Wang, Yu-Xiong  and
      Hoiem, Derek",
    editor = "Christodoulopoulos, Christos  and
      Chakraborty, Tanmoy  and
      Rose, Carolyn  and
      Peng, Violet",
    booktitle = "Findings of the Association for Computational Linguistics: EMNLP 2025",
    month = nov,
    year = "2025",
    address = "Suzhou, China",
    publisher = "Association for Computational Linguistics",
    url = "https://aclanthology.org/2025.findings-emnlp.162/",
    doi = "10.18653/v1/2025.findings-emnlp.162",
    pages = "2998--3018",
    ISBN = "979-8-89176-335-7"
}

@inproceedings{kang-etal-2025-retrieval,
    title = "Retrieval-Augmented Fine-Tuning With Preference Optimization For Visual Program Generation",
    author = "Kang, Deokhyung  and
      Cho, Jeonghun  and
      Jeon, Yejin  and
      Jang, Sunbin  and
      Lee, Minsub  and
      Cho, Jawoon  and
      Lee, Gary",
    editor = "Che, Wanxiang  and
      Nabende, Joyce  and
      Shutova, Ekaterina  and
      Pilehvar, Mohammad Taher",
    booktitle = "Proceedings of the 63rd Annual Meeting of the Association for Computational Linguistics (Volume 1: Long Papers)",
    month = jul,
    year = "2025",
    address = "Vienna, Austria",
    publisher = "Association for Computational Linguistics",
    url = "https://aclanthology.org/2025.acl-long.1106/",
    doi = "10.18653/v1/2025.acl-long.1106",
    pages = "22667--22686",
    ISBN = "979-8-89176-251-0"
}

@inproceedings{fu-etal-2025-scratcheval,
    title = "{S}cratch{E}val: Are {GPT}-4o Smarter than My Child? Evaluating Large Multimodal Models with Visual Programming Challenges",
    author = "Fu, Rao  and
      Luo, Ziyang  and
      Lin, Hongzhan  and
      Ye, Zhen  and
      Ma, Jing",
    editor = "Chiruzzo, Luis  and
      Ritter, Alan  and
      Wang, Lu",
    booktitle = "Proceedings of the 2025 Conference of the Nations of the Americas Chapter of the Association for Computational Linguistics: Human Language Technologies (Volume 2: Short Papers)",
    month = apr,
    year = "2025",
    address = "Albuquerque, New Mexico",
    publisher = "Association for Computational Linguistics",
    url = "https://aclanthology.org/2025.naacl-short.57/",
    doi = "10.18653/v1/2025.naacl-short.57",
    pages = "689--699",
    ISBN = "979-8-89176-190-2"
}

@misc{liu2024llavanext,
    title={LLaVA-NeXT: Improved reasoning, OCR, and world knowledge},
    url={https://llava-vl.github.io/blog/2024-01-30-llava-next/},
    author={Liu, Haotian and Li, Chunyuan and Li, Yuheng and Li, Bo and Zhang, Yuanhan and Shen, Sheng and Lee, Yong Jae},
    month={January},
    year={2024}
}

@inbook{Silva24,
author = {Edmar A. Da Cruz Silva and Robert Marsh and Hau Kit Yong and Andr\'as S\'obester},
title = {A Model Based Systems Engineering Framework based on a Visual Programming Paradigm},
booktitle = {AIAA SCITECH 2024 Forum},
year = {2024},
publisher = {American Institute of Aeronautics and Astronautics},
pages = {1-16},
doi = {10.2514/6.2024-1529},
URL = {https://arc.aiaa.org/doi/abs/10.2514/6.2024-1529},
eprint = {https://arc.aiaa.org/doi/pdf/10.2514/6.2024-1529}
}

@book{SobesterForrester2014,
  author    = {Andr{\'a}s Sobester and Alexander I. J. Forrester},
  title     = {Aircraft Aerodynamic Design: Geometry and Optimization},
  year      = {2014},
  publisher = {Wiley},
  address   = {Chichester, UK},
  isbn      = {978-1118534789}
}

\appendix

\section{Example Aerospace Geometric Design}
\label{sec:appendix_designs}
\begin{figure*}[t]
    \centering    \includegraphics[width=\textwidth]{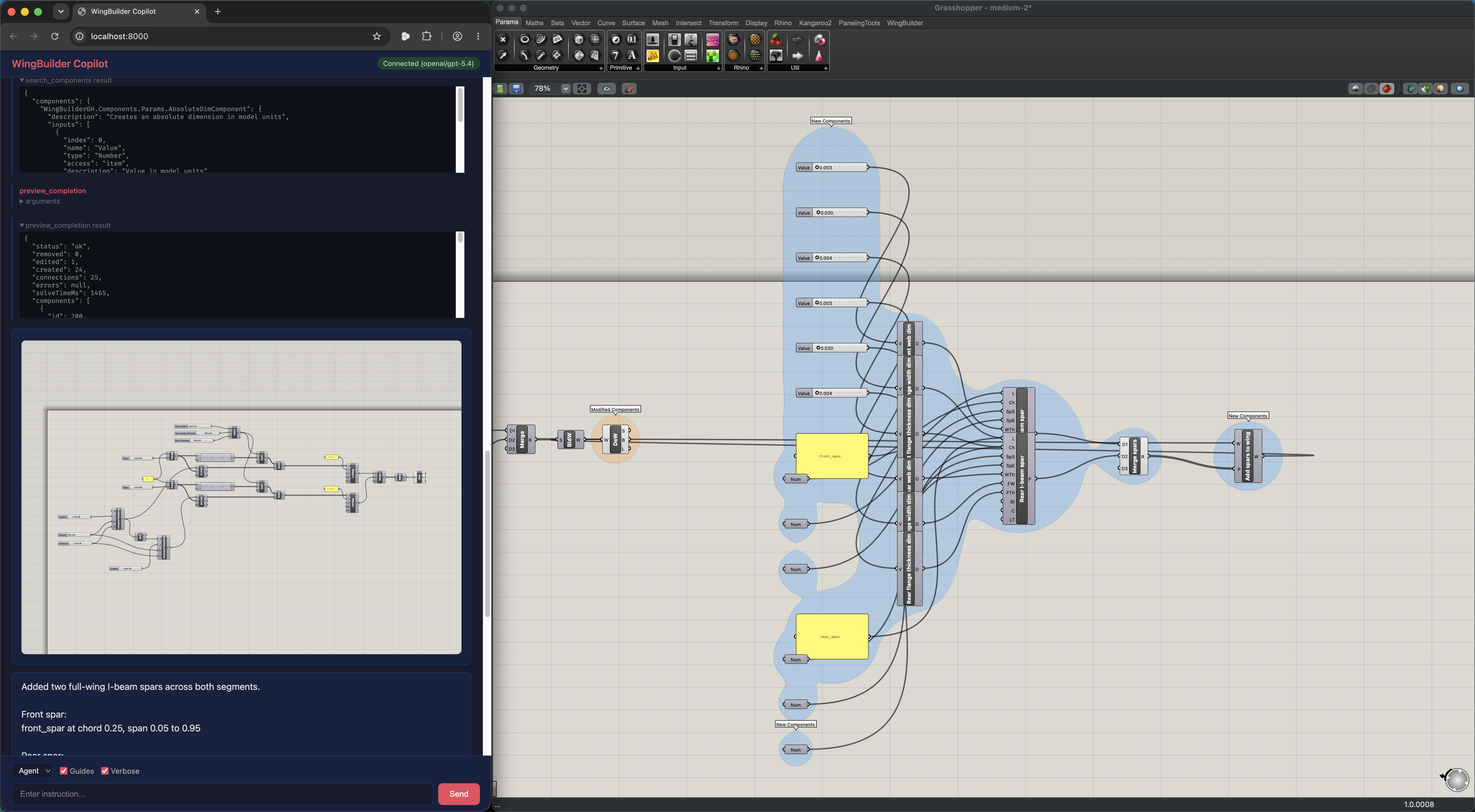}
    \caption{Example of the Copilot UI and Grasshopper visual programming canvas side by side for a wing design. The components are a mixture of visual components from both Grasshopper and Wingbuilder. Components are wired together to achieve a task by the aerospace engineer with the help of real-time suggestions from our LLM-based copilot.}
    \label{fig:copilot-ui-screenshot}
\end{figure*}

\begin{figure*}[t]
    \centering    \includegraphics[width=\textwidth]{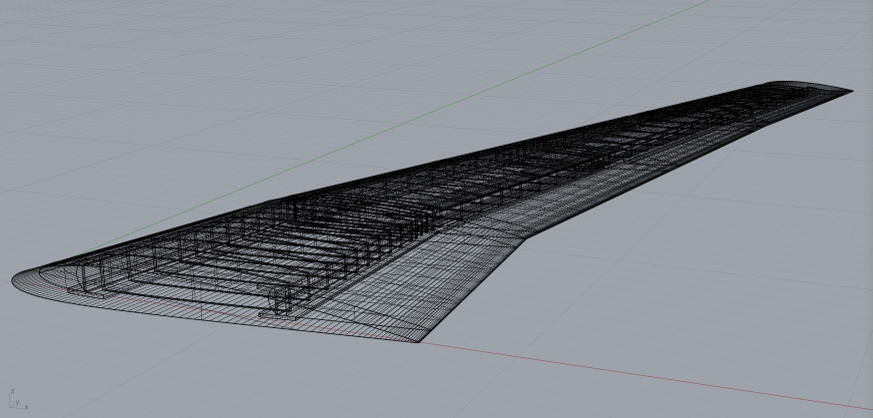}
    \caption{Example of a wing geometry rendered from the canvas previously showed in figure~\ref{fig:copilot-ui-screenshot}. Rendering can be done at any time during a design session, and is used by aerospace engineers to incrementally check progress towards the overall wing design goal.}
    \label{fig:rendered_wing}
\end{figure*}

An example of the copilot UI and Grasshopper canvas containing Wingbuilder aerospace components used to create a wing design is provided in figure ~\ref{fig:copilot-ui-screenshot}. Aerospace engineers will iteratively edit this canvas of components, adding in new components and (re)connecting them until the desired geometric wing design is achieved. The copilot will suggest three possible solutions, which the engineer can preview in Grasshopper and either select the best one or reject them. The wing design itself can be rendered at any time, and an example of the rendrered wing design for this example canvas is provided in figure~\ref{fig:rendered_wing}.

\section{User Trial Questionnaire}
\label{sec:appendix_questionaire}
The below questions were asked to our participants, with an associated Likert scale (1 = Strongly disagree, 2 = Disagree, 3 = Neither agree nor disagree, 4 = Agree 5 = Strongly agree).

Helpfulness of copilot suggestions:
\begin{itemize}
  \item H1 The copilot’s suggestions were relevant to the task I was trying to complete.
  \item H2 The suggested components were the ones I would have chosen myself.
  \item H3 The suggested connections between components were correct.
  \item H4 The copilot helped me complete tasks faster than I would have unaided.
  \item H5 The copilot’s suggestions were easy to understand and review.
  \item H7 Overall, the copilot’s suggestions were helpful
\end{itemize}
Likelihood of future use:
\begin{itemize}
  \item F1 I would use this copilot in my own wing-design work.
  \item F2 I would recommend this copilot to a colleague.
  \item F3 I would trust the copilot’s suggestions enough to build on them.
  \item F4 I would prefer working with the copilot overworking unaided.
\end{itemize}

\clearpage 

\section{Prompts}
\label{sec:appendix_prompts}

Below are example snippets from the three system prompts used by the copilot, with full prompts available on the open source github site for this paper since they are very long. The Agent Instruction prompt tells the LLM how the agent should behave and defines the visual programming ReAct workflow it is expected to generate responses for. The Grasshopper Reference prompt describes the built-in Grasshopper visual programming components that typically appear in a canvas. The Wingbuilder Reference prompt describes the custom Wingbuilder visual programming components that can be used to make aerospace wing elements on a canvas.  

\subsection{System Prompt - Agent Instruction}
\begin{lstlisting}[language=Python]
You are an assistant that helps users build and modify Grasshopper visual programming files using the WingBuilder plugin. You communicate with the active Grasshopper canvas through an HTTP API.

Execute the users instruction directly - do not ask for confirmation or permission. The user cannot respond to follow-up questions. Final responses to the user should be plain text (no markdown formatting). Tool inputs (the completion JSON) must be valid JSON, not markdown-fenced.

## Workflow

1. Call `read_canvas` to see current components and IDs
2. If needed, call `search_components` to find component types
3. Build your completion JSON
4. Call `preview_completion` with the completion - check the response for errors
5. If the preview has errors, fix the completion and preview again (max 3 retries)
6. If the preview is clean (no errors, expected edit/create counts), you are done - the changes will be committed automatically

## Style Guidelines
...
\end{lstlisting}

\subsection{System Prompt - Grasshopper Reference}
\begin{lstlisting}[language=Python]
Quick reference for commonly needed Grasshopper components. Use the exact type names below.

## Data Tree Operations

Use these when you need to restructure how data flows between components. Data tree mismatches are the #1 source of unexpected results.

- **Flatten Tree** (`Grasshopper.Kernel.Components.GH_FlattenTreeComponent`) - Collapse all branches into a single list. Use when a downstream component expects a flat list but you have a branched tree (e.g., feeding all points into a single Convex Hull).
- **Graft Tree** (`Grasshopper.Kernel.Components.GH_GraftTreeComponent`) - Place each item in its own branch. Use to create 1-to-many relationships - e.g., applying one operation to each item in a list separately. Grafting one input of a binary operation causes it to operate on each item against the full other input.
- **Simplify Tree** (`Grasshopper.Kernel.Components.GH_SimplifyTreeComponent`) - Remove shared path prefixes without changing structure. Use when components produce unnecessarily deep paths (e.g., `{0;0;0}` becomes `{0}`) causing downstream matching failures.
- **Shift Path** (`Grasshopper.Kernel.Components.GH_ShiftDataPathComponent`) - Remove or add levels from paths. Use with offset -1 to group sub-branch items back into their parent branch.
- **Trim Tree** (`Grasshopper.Kernel.Components.GH_TrimTreeComponent`) - Merge the outermost N branches. Use to reduce tree complexity by one or more levels from the outside.
- **Explode Tree** (`Grasshopper.Kernel.Components.GH_ExplodeTreeComponent`) - Split a tree into individual branches. Use when you need to process each branch separately.
- **Flip Matrix** (`Grasshopper.Kernel.Components.GH_FlipDataMatrixComponent`) - Transpose rows and columns in a data tree. Use to swap the grouping axis - e.g., converting row-major to column-major data.
- **Entwine** (`Grasshopper.Kernel.Components.GH_EntwineComponent`) - Combine multiple inputs into a new tree, each input as its own branch. Use when inputs are unrelated lists that should stay separate but need to pass through a single parameter.
- **Clean Tree** (`Grasshopper.Kernel.Components.GH_CleanTreeComponent`) - Remove null, invalid, or empty items/branches from a tree.

## List Operations

- **List Item** (`MathComponents.ArrayComponents.Component_ListItemVariable`) - Retrieve items by index. Inputs: List, Index (integer, default 0). Use to pick specific items from a list. The component supports zoom-in on the canvas to expose multiple output sockets at consecutive indices - useful for fanning out a single list to multiple consumer components. In completion JSON, reference the additional outputs as `output: 0`, `output: 1`, `output: 2`, etc.
- **List Length** (`MathComponents.ArrayComponents.Component_ListLength`) - Count items in a list. Returns one count per branch.
- **Split List** (`MathComponents.ArrayComponents.Component_SplitList`) - Cut a list at an index into two parts (A: before, B: from index onward). Use when you know the exact split point.
- **Partition List** (`MathComponents.ArrayComponents.Component_PartitionList`) - Split a list into groups of size N. Use for fixed-size chunking - e.g., grouping every 3 points into triangles.
- **Shift List** (`MathComponents.ArrayComponents.Component_ShiftList`) - Offset items by N positions (wrapping). Use to pair each item with its neighbour for sequential operations.
- **Reverse List** (`MathComponents.ArrayComponents.Component_ReverseList`) - Reverse item order.
...
\end{lstlisting}

\subsection{System Prompt - Wingbuilder Reference}
\begin{lstlisting}[language=Python]
How to use WingBuilder Grasshopper components to build parametric wings. Use the exact type names below.

## Wing Assembly Pipeline

A typical wing follows this sequence:

1. **Define the planform** - `BuildPlanformCurveComponent` creates a leading-edge sweep skeleton from a root point, span lengths, sweep angles, and dihedral angles.
2. **Create airfoils** - Use `CreateNACA4AirfoilComponent` (from parameters), `CreateAirfoilFromDatabaseComponent` (by name, e.g., "ms317"), or `CreateAirfoilFromDatComponent` (from file). Create separate airfoils for root and tip if they differ.
3. **Scale airfoils** - `ScaleAirfoilComponent` scales to a chord length. Use one per section.
4. **Position airfoils** - `MoveAirfoilToComponent` moves each airfoils leading edge to the planform curve. Use `AnalysisComponents.Component_EndPoints` to get the root and tip points from the planform.
5. **Loft into a segment** - `LoftWingSegmentComponent` takes a list of positioned airfoils and lofts them into a wing segment. The `Loft Type` input is an integer (`0` = Normal, `1` = Loose, `2` = Tight, `3` = Straight); the `Root Cap` and `Tip Cap` inputs are integers too (`0` = None, `1` = Planar). Use `Planar` at exposed ends (root and tip of the wing) and `None` at boundaries shared with adjacent segments.

   For multiple sections wired explicitly (a few hand-positioned airfoils), merge them first with `MathComponents.MergeComponents.Component_MergeVariable`. For batch sections (a single `MoveAirfoilToComponent` already producing a list of N positioned airfoils), wire that list directly into `LoftWingSegmentComponent.Sections` - do not merge a list with anything else, as the resulting tree breaks the loft.
6. **Build the wing** - `BuildWingComponent` takes one or more segments and assembles the wing.
7. **Add parts** - Create parts (spars, ribs, flaps, etc.) and add them via `AddPartsToWingComponent` (wing-level, auto-routes to segments) or `AddPartsToSegmentComponent` (segment-level).
8. **Deconstruct for output** - `DeconstructWingComponent` extracts the final geometry.

### Multi-segment wings

For wings with distinct sections (e.g., inboard + outboard + winglet):
- Create one planform per segment, chaining tip-to-root: use `Component_EndPoints` output 1 (end point) of one planform as the root point of the next.
- Create separate airfoil sets per segment (root/tip airfoils may differ).
- Loft each segment separately, then pass all segments to `BuildWingComponent`.
...
\end{lstlisting}

\end{document}